\documentclass[12pt]{style}
\usepackage[toc,page,header]{appendix}
\usepackage{wrapfig}

\usepackage{natbib}

\usepackage{CJKutf8}

\usepackage{xargs}

\usepackage{todonotes}

\usepackage{multirow}

\usepackage{cleveref}

\usepackage{amsmath}
\usepackage{amssymb}
\usepackage{mathtools}
\usepackage{amsthm}
\usepackage{dsfont}

\usepackage{svg}

\usepackage{mathrsfs}
\usepackage{adjustbox}
\usepackage{multicol}
\usepackage{tcolorbox}
\usepackage{changepage}
\usepackage{enumitem}
\usepackage{graphicx}
\usepackage{xcolor}
\usepackage{float}
\usepackage{threeparttable}
\usepackage{subcaption}
\usepackage{algorithm}
\usepackage{algpseudocode}
\usepackage{wrapfig}
\usepackage[table]{xcolor}
\usepackage{colortbl}
\usepackage{tabularx}
\usepackage{makecell}
\usepackage[dvipsnames]{xcolor}

\newcolumntype{g}{>{\columncolor{gray!10}}c}

\theoremstyle{plain}

\theoremstyle{definition}

\theoremstyle{remark}

\definecolor{catgray}{gray}{0.9}
\definecolor{skyblue}{rgb}{0.53,0.81,0.92}
\colorlet{skyblue!30}{skyblue!30!white}
\definecolor{customblue}{RGB}{70,130,180}

\renewcommand{\emph}[1]{\textit{#1}}

\usepackage{minitoc}
\usepackage{url}

\PassOptionsToPackage{table,xcdraw}{xcolor}
\usepackage{titletoc}
\usepackage{placeins}
\usepackage{pifont}

\definecolor{RowBlue}{HTML}{E9F2FB}
\definecolor{RowRed}{HTML}{F9EAEA}
\definecolor{Top1}{HTML}{50DB4B}
\definecolor{Top2}{HTML}{A5FFA2}
\definecolor{Top3}{HTML}{D9FFD9}
\definecolor{Sub1}{HTML}{EAB8B8}
\definecolor{Sub2}{HTML}{E4E4E4}

\title{MiniWorld: Democratizing the Training of Video World Models from Scratch}

\author[1]{Yian Zhao}
\author[1]{Ruochong Zheng}
\author{Hongcan Guo}
\author[1]{Yu Yan}
\author[1]{Jian Zhang}
\author[1]{Jie Chen}
\affiliation[1]{Peking University}

\newcommand{\githubicon}{\raisebox{-0.15em}{\includegraphics[height=1.05em]{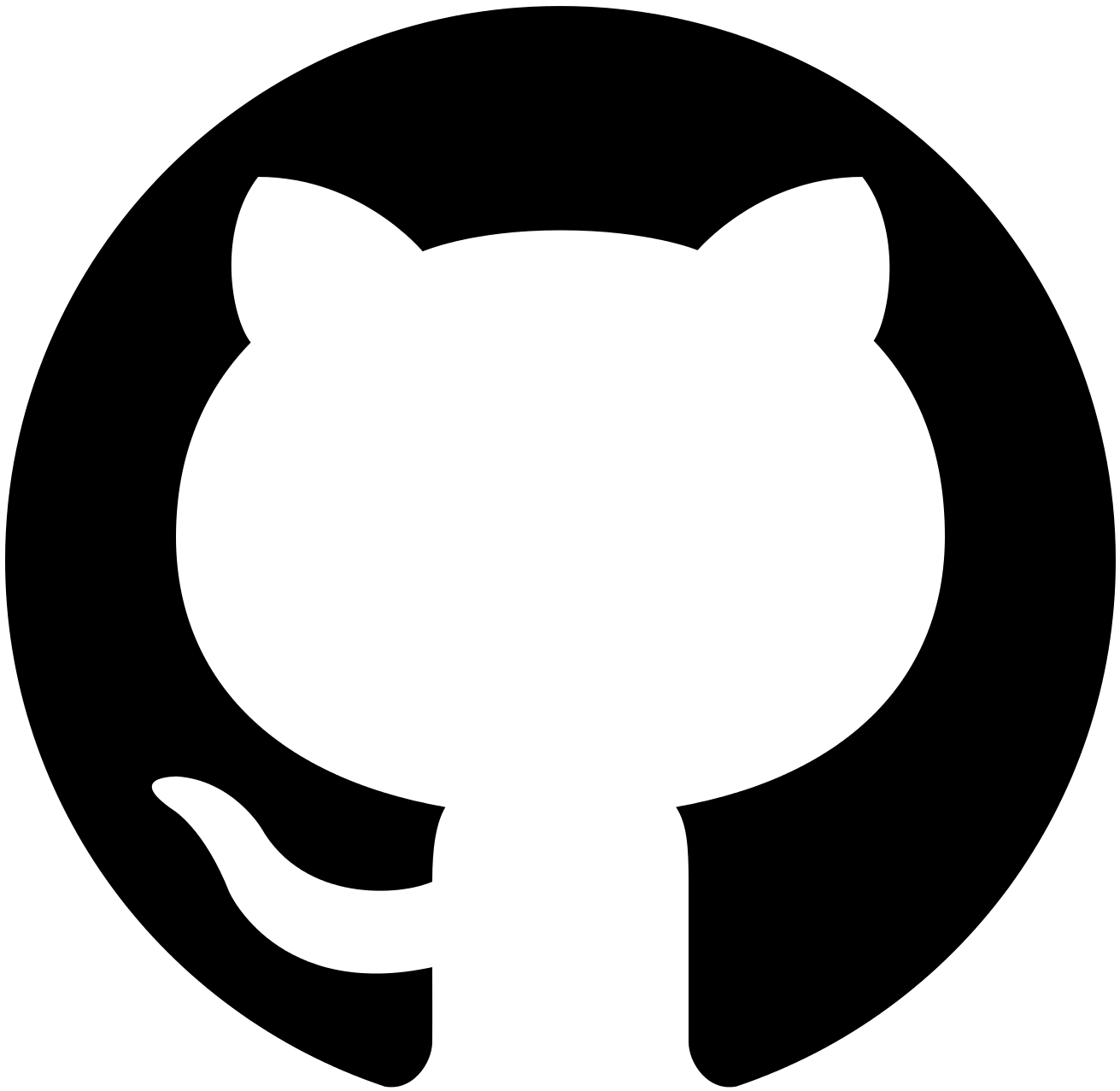}}}
\newcommand{\hficon}{\raisebox{-0.15em}{\includegraphics[height=1.05em]{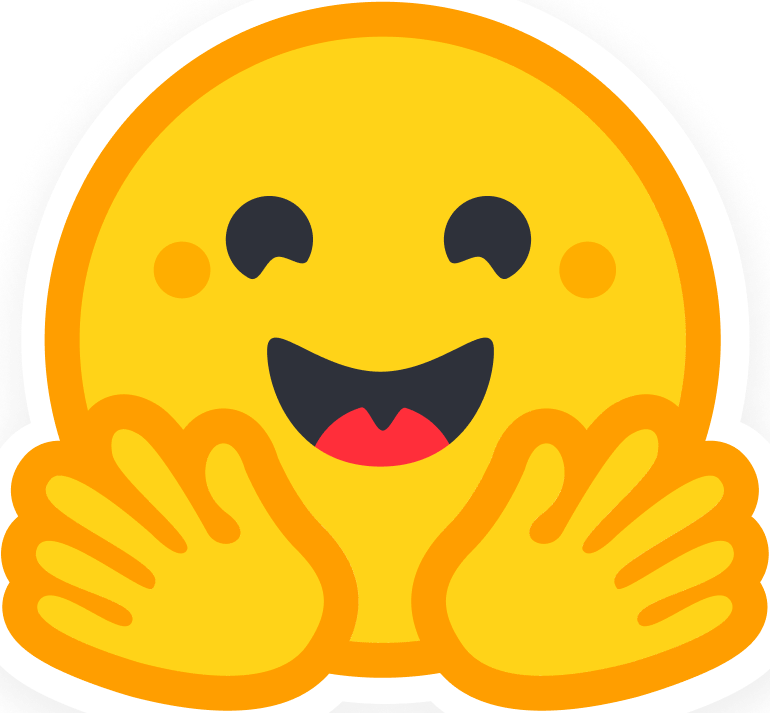}}}
\newcommand{\globeicon}{\raisebox{-0.15em}{\includegraphics[height=1.05em]{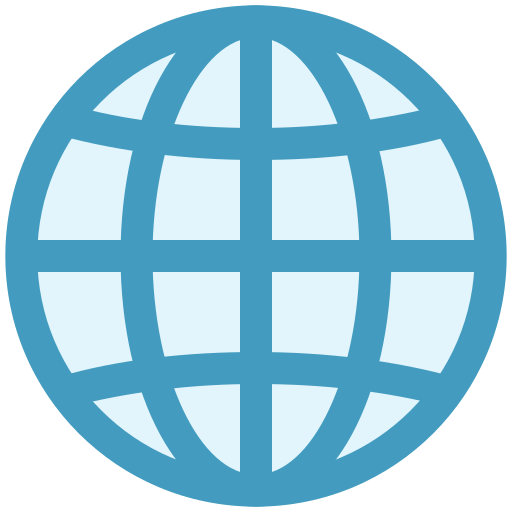}}}
\newcommand{\resourcelink}[1]{\href{#1}{\texttt{#1}}}

\abstract{
Video world models predict future observations conditioned on historical observations and control signals, enabling long-horizon generation through autoregressive state transitions.
Unlike conventional video generation models that primarily capture visual appearance and motion, video world models learn the underlying dynamics governing environment evolution under agent actions.
By modeling how the world responds to interactions, they provide a fundamental building block for embodied AI and interactive simulation.
Recent progress has largely relied on adapting pretrained video generation models through post-training or distillation. Although effective, these approaches often require complex training pipelines, substantial computational resources, and inevitably suffer from the mismatch between bidirectional pretraining and causal streaming inference.
Recent studies have shown that training autoregressive video world models from scratch is both feasible and scalable. 
However, the community still lacks a lightweight, transparent, and fully reproducible baseline that can be trained end-to-end with modest computational resources.
We present MiniWorld, an accessible and reproducible framework for training streaming video world models from scratch. MiniWorld employs a block-causal Video Diffusion Transformer trained with Flow Matching in the latent space of a pretrained Video VAE.
Building on Diffusion Forcing, it adopts a chunk-wise non-decreasing noise schedule together with two-stage continued training to improve temporal modeling and training stability.
During inference, MiniWorld combines a rolling KV cache with pipelined asynchronous denoising for efficient streaming generation under bounded computation.
The entire model can be trained within several days on a single 8-GPU server.
By releasing the complete training and inference codebase, pretrained model checkpoints, and implementation details, we hope MiniWorld will facilitate future research on video world modeling.
}

\checkdata[\githubicon\hspace{0.45em}Code]{\resourcelink{https://github.com/zhao-yian/MiniWorld}}
\checkdata[\hficon\hspace{0.45em}Models]{\resourcelink{https://huggingface.co/zhaoyian01/MiniWorld}}
\checkdata[\globeicon\hspace{0.45em}Project Page]{\resourcelink{https://zhao-yian.github.io/MiniWorld}}

\begin{document}
\maketitle

\section{Introduction}

Video world models have recently emerged as a promising paradigm for interactive visual simulation, where future observations are generated autoregressively from historical observations and control signals. Unlike conventional video generators that primarily synthesize realistic appearance and motion, video world models aim to capture the dynamics governing world evolution, enabling persistent state prediction over long horizons. Such capabilities are becoming increasingly important for embodied AI, interactive environment simulation, and general physical intelligence. Recent systems, including Genie 3~\citep{genie3}, Cosmos~\citep{cosmos}, LingBot-World~\citep{lingbotworld}, DreamX-World 1.0~\citep{dreamxworld}, Matrix-Game 3.0~\citep{matrixgame3}, HappyOyster~\citep{happyoyster}, DreamDojo~\citep{dreamdojo}, MAGI-1~\citep{magi}, and SkyReels-V2~\citep{skyreelsv2}, have demonstrated the remarkable potential of large-scale video world models across these applications.

Despite this rapid progress, training a video world model from scratch remains prohibitively expensive for most researchers. The prevailing paradigm adapts pretrained bidirectional video diffusion models, such as Wan~\citep{wan}, into autoregressive world models through post-training techniques including fine-tuning, distillation, and reinforcement learning. While this approach effectively transfers the strong visual generation capability of large video foundation models, it requires complex multi-stage optimization pipelines and substantial computational resources. More fundamentally, bidirectional pretraining is intrinsically inconsistent with causal streaming inference, resulting in a structural mismatch between training and deployment.

Existing work has demonstrated the feasibility of end-to-end autoregressive pretraining for video world models. MAGI-1~\citep{magi} demonstrates the scalability of chunk-wise autoregressive video pretraining, while SkyReels-V2~\citep{skyreelsv2} further shows that Diffusion Forcing~\citep{df} provides an effective formulation for long-horizon video generation and world modeling. These advances suggest that training streaming video world models directly from scratch is becoming increasingly practical. Nevertheless, the community still lacks a lightweight and fully reproducible baseline that prioritizes simplicity, transparency, and accessibility over the continuous expansion of model scale and improvement of in-house data quality.

In this work, we present MiniWorld, an accessible recipe for training streaming video world models from scratch under modest computational budgets. MiniWorld adopts a block-causal Video Diffusion Transformer trained with Flow Matching in latent space. Building upon Diffusion Forcing, we introduce a chunk-wise non-decreasing noise scheduling strategy and a two-stage continued training paradigm, enabling efficient and stable world modeling while preserving a simple training procedure.

During inference, MiniWorld employs a rolling KV cache with pipelined asynchronous denoising, allowing long-horizon world generation under bounded computation and a flexible trade-off between generation quality and inference throughput. Unlike many recently released systems, MiniWorld deliberately avoids introducing proprietary datasets or sophisticated post-training. Instead, it delivers an essential minimal end-to-end pipeline covering data preprocessing, model training, streaming inference, and evaluation. The complete model can be trained in several days on a single 8-GPU server, making streaming world model research substantially more accessible.

Our contributions are summarized as follows:
\begin{itemize}[leftmargin=*]
  \item We introduce MiniWorld, a lightweight and fully reproducible framework that makes training streaming video world models from scratch feasible under modest computational budgets.
  \item We develop a practical streaming world modeling recipe that integrates block-causal Video DiT, chunk-wise non-decreasing noise scheduling, two-stage continued training, and rolling-KV streaming inference.
  \item We release the full data processing, training, inference, and evaluation pipeline, offering a transparent and extensible open-source baseline for future research.
\end{itemize}
We hope MiniWorld demonstrates that stable long-horizon, action-conditioned streaming world modeling does not require prohibitive compute, and can instead be studied through a compact, open, and reproducible experimental platform.

\section{Preliminaries: Rectified Flow}

MiniWorld is built upon Rectified Flow~\citep{rf}, a simple formulation of Flow Matching~\citep{fm}. Following modern latent video generation frameworks, all modeling is performed in the latent space of a pretrained video VAE. Given a latent video sample \(x\) and Gaussian noise \(\epsilon\), Rectified Flow constructs a linear interpolation
\begin{equation}
z_{\tau} = (1-\tau)x + \tau\epsilon,\qquad \tau\in[0,1],
\end{equation}
and trains the model to predict the corresponding velocity field \(v^{*}(x,\epsilon)=x-\epsilon\).

During inference, generation starts from Gaussian noise and progressively integrates the learned velocity field to recover the target latent sequence. Throughout this paper, we use the pretrained Wan2.2 VAE to encode videos into latent representations, and all training and streaming inference are conducted in this latent space.

\begin{figure}[!t]
  \centering
  \includegraphics[width=\linewidth]{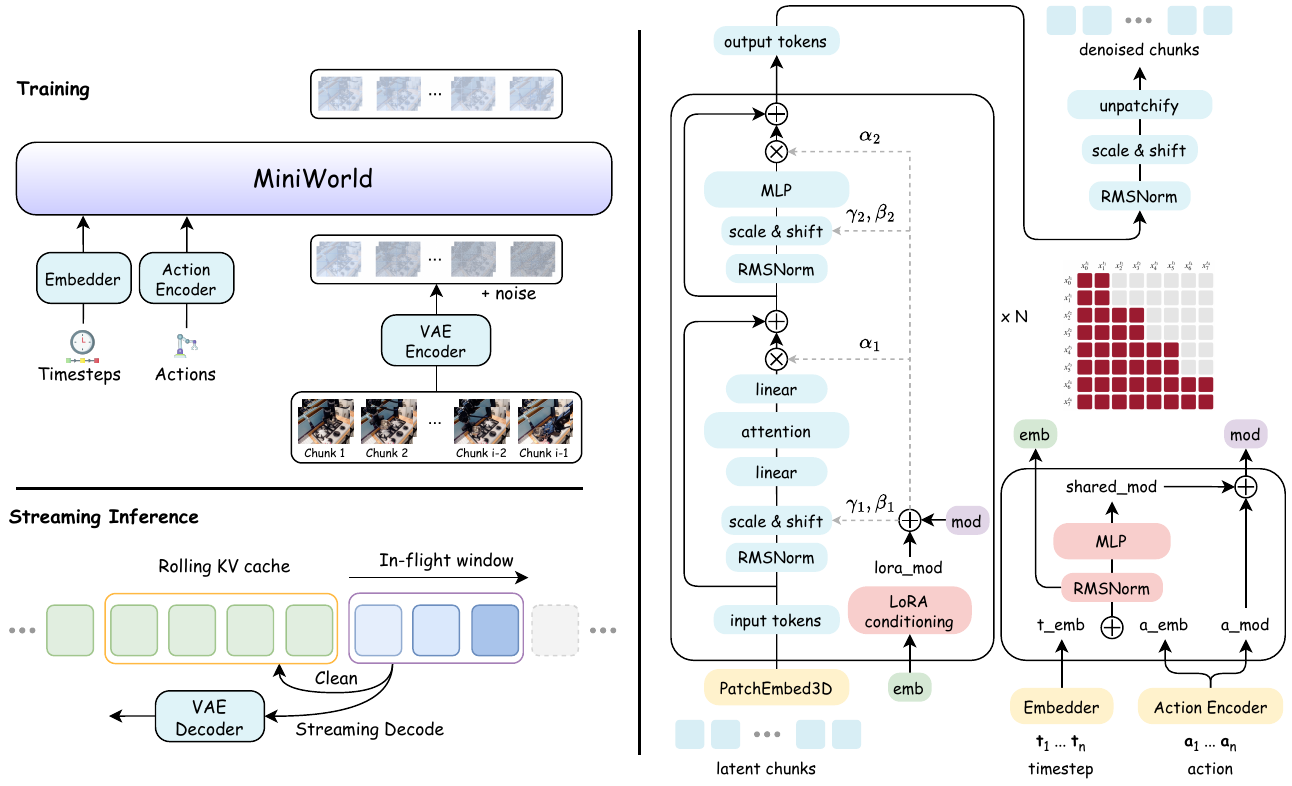}
  \caption{Overview of MiniWorld. Videos are encoded into latent representations by a pretrained Video VAE, while actions are converted into latent-frame-aligned conditioning signals. MiniWorld partitions the latent sequence into temporal chunks and trains a block-causal Video DiT with chunk-wise non-decreasing noise schedules. During streaming inference, future chunks are denoised asynchronously, completed chunks are committed to a rolling KV cache as persistent history, and the active denoising window remains bounded as generation proceeds.}
  \label{fig:method}
\end{figure}

\section{Method}
\label{sec:method}

In this section, we present the proposed MiniWorld framework, illustrated in Figure~\ref{fig:method}. We first formulate streaming video world modeling as a next-state prediction problem, and then describe the block-causal Video DiT architecture, conditioning mechanism, training strategy, and streaming inference pipeline.

\subsection{Problem Formulation}

The objective of a video world model is to predict future observations conditioned on historical observations and control signals. Formally, given a sequence of observed frames \(x_{1:h}\) and future actions \(c_{h+1:T}\), the model aims to generate the future video sequence \(x_{h+1:T}\).

A straightforward solution is to formulate this task as conditional video generation within a fixed temporal window. During training, historical frames remain clean while future frames are jointly perturbed by diffusion noise, allowing a bidirectional Video DiT to denoise all future frames in parallel. Since every predicted frame can attend to all other future frames, this formulation fully exploits future temporal context and has demonstrated strong performance on short-video generation and video completion.

However, long-horizon deployment typically requires autoregressive streaming inference. A common strategy is to repeatedly generate a fixed-length video segment, append the generated frames to the history, and slide the temporal window forward. This introduces a fundamental mismatch between training and inference. During training, the historical context always comes from ground-truth videos, whereas during inference it is replaced by previously generated predictions. Consequently, prediction errors accumulate over time and continuously propagate to future generations, resulting in the well-known exposure bias. Furthermore, because the temporal window has a fixed length, early observations are progressively discarded as generation proceeds. Without an explicit mechanism for maintaining long-term world states across windows, the model gradually loses historical information, leading to temporal drift and degraded long-range consistency.

Despite these limitations, the continuation paradigm remains attractive because it can directly leverage the strong visual generation capability of existing bidirectional video diffusion models. Accordingly, recent studies have sought to improve this framework from different perspectives, including reducing the training--inference mismatch, introducing causal modeling, and maintaining rolling historical states through KV caching~\citep{rolling,causalforcing,selfforcing}.

The above analysis suggests that the primary bottleneck of fixed-window video diffusion lies not in visual generation quality itself, but in the inherent discrepancy between the training objective and the streaming inference process. Rather than adapting a bidirectional video diffusion model through post hoc modifications, MiniWorld is built directly upon an action-conditioned next-state prediction formulation. This formulation substantially reduces the mismatch between training and inference while naturally supporting long-horizon streaming generation and persistent world-state transitions.

\subsection{Block-Causal Video DiT}

MiniWorld is built upon a single-stream Video Diffusion Transformer (Video DiT) trained with Rectified Flow~\citep{rf}. Given a noisy latent video, a 3D patch embedding layer converts the latent volume into visual tokens. Diffusion timesteps, robot actions, and camera poses are incorporated as conditioning signals through a unified modulation framework.

To enable streaming generation, MiniWorld employs block-causal self-attention, which partitions the video sequence into a series of temporally ordered chunks. Tokens within the same chunk attend bidirectionally to each other, while cross-chunk attention is strictly causal, such that each chunk can only access preceding chunks. This attention pattern preserves rich spatio-temporal interactions within individual chunks while enforcing the causal dependency required for autoregressive streaming generation.

\noindent{\sffamily\bfseries Action Conditioning.}
Robot actions are incorporated through an AdaLN-based conditioning mechanism. For each latent frame, the action encoder transforms the action input \(a_t\) into an action embedding \(e_t^a\) and an action modulation vector \(m_t^a\):
\begin{equation}
(e_t^a,m_t^a)=E_{\mathrm{act}}(a_t).
\end{equation}
The action embedding is added to the diffusion timestep embedding and passed through a shared modulation network to produce the base AdaLN parameters. The action modulation vector is then added residually to the generated parameters before being applied to all Transformer blocks.

This design explicitly decouples semantic conditioning from feature modulation. The action embedding interacts with the diffusion timestep to capture the global denoising state, while the modulation vector directly adjusts the AdaLN parameters, providing a more expressive and efficient mechanism for action conditioning.

To further improve conditioning flexibility, we introduce an AdaLN-LoRA modulation network. Instead of learning an independent modulation network for every Transformer block, all layers share a common modulation branch, while each block learns only a lightweight low-rank residual,
\begin{equation}
M_{\ell}(e)=M_{\mathrm{shared}}(e)+W^{\ell}_{\mathrm{up}}\,
\sigma\!\left(W^{\ell}_{\mathrm{down}}e\right).
\end{equation}
The low-rank residual is zero-initialized so that the model starts from a fully shared modulation network and progressively learns layer-specific adaptations during training, increasing conditioning capacity with only a marginal parameter overhead.

During training, structured condition dropout is performed by replacing action inputs with a null condition for classifier-free guidance. Since the initial observation has no preceding action, the first latent frame is always assigned the null action embedding.

\noindent{\sffamily\bfseries Model Scaling.}
MiniWorld adopts a unified architecture family with model sizes up to 3B parameters. All variants share the same block-causal Transformer architecture, conditioning interface, and patch size, differing only in depth, width, and number of attention heads. Unless otherwise specified, we report results using two configurations: MiniWorld-0.5B (28 layers, hidden size 1152, 16 attention heads) and MiniWorld-1B (28 layers, hidden size 1536, 12 attention heads).

\subsection{Chunk-wise Noise Scheduling}

Following Diffusion Forcing~\citep{df}, MiniWorld assigns independent diffusion timesteps to different latent chunks within the same training sequence, rather than conditioning each prediction on a fully denoised previous chunk as in teacher-forcing-based autoregressive diffusion. This formulation avoids explicitly unrolling next-chunk prediction during training and enables asynchronous denoising across chunks, allowing multiple future chunks to be processed concurrently during streaming inference.

We further impose a non-decreasing timestep constraint on chunk-wise noise levels,
\begin{equation}
\tau_1\le\tau_2\le\cdots\le\tau_M,
\end{equation}
where \(\tau=0\) denotes the clean data and \(\tau=1\) denotes pure noise. This constraint enforces a causal denoising order in which earlier chunks are always cleaner than later ones, reflecting the practical inference state that past observations have largely converged while future contents remain uncertain. As discussed in AR-Diffusion~\citep{ardiff}, restricting timestep compositions to monotonic schedules also substantially reduces the search space of asynchronous diffusion trajectories, leading to more stable optimization and faster convergence.

Following the probability propagation strategy in AR-Diffusion~\citep{ardiff}, we extend the frame-level Frame-oriented Probability Propagation (FoPP) scheduler to chunk-wise generation, resulting in a Chunk-oriented Probability Propagation (CoPP) scheduler. CoPP maintains the balanced sampling property of FoPP over both timestep compositions and individual diffusion timesteps while operating on latent chunks. Specifically, an anchor chunk is first randomly selected and assigned a timestep sampled from a logit-normal distribution. The timesteps of neighboring chunks are then propagated outward from the anchor under the non-decreasing constraint, producing a valid chunk-wise timestep composition.

Compared with naively sampling monotonic timestep sequences, CoPP preserves the favorable timestep distribution of FoPP while exposing the model to a diverse set of valid chunk-wise diffusion trajectories, improving robustness across different streaming inference schedules.

Given the sampled timestep sequence, flow matching is performed only on the prediction chunk set \(\Omega\), while observed history chunks are treated as conditions and excluded from the optimization:
\begin{equation}
\mathcal{L}_{\mathrm{FM}}=
\frac{1}{|\Omega|}\sum_{n\in\Omega}
\left\|
\left[v_{\theta}(z_{\tau},\tau,c)\right]_n-(x_n-\epsilon_n)
\right\|_2^2.
\end{equation}

To support different streaming generation scenarios within a unified training framework, MiniWorld randomly adopts one of two in-context conditioning modes during training. In image-to-video mode, only the first frame is kept clean, while the remaining frames in the first chunk are diffused. In video-to-video mode, the entire first chunk is treated as observed context and remains noise-free. Randomly mixing the two conditioning modes enables the model to seamlessly support both image-to-video and video-to-video generation at inference time without changing the training objective.

\subsection{Two-stage Continued Training}

To improve long-horizon generation while maintaining training efficiency, MiniWorld adopts a two-stage training strategy. In the pre-training stage, the model is trained on short video clips of 21/46 frames, allowing it to efficiently learn local action-conditioned state transitions with substantially reduced computational cost. The model is then continually trained on longer sequences of 125/253 frames at the same spatial resolution. Rather than modifying the training objective, the second stage simply exposes the model to longer causal contexts, longer action trajectories, and longer camera-pose trajectories.

We apply timestep shifting during the long-horizon training stage by reparameterizing the diffusion timestep as
\begin{equation}
\tilde{\tau}=\frac{s\tau}{1+(s-1)\tau},
\end{equation}
where \(s\) denotes the shift factor. Timestep shifting reallocates the sampling density over diffusion timesteps and has been shown to improve optimization for large-scale diffusion models~\citep{sd3}. The long-horizon training stage exposes the model to substantially longer temporal contexts, enabling it to better exploit extended histories during inference.

\subsection{Streaming Inference}

At inference time, MiniWorld performs streaming generation through asynchronous chunk-wise denoising. At any moment, the sequence consists of two parts: a committed history that has already converged to the data manifold and a set of future chunks that are still undergoing denoising. During each denoising step, a future chunk is allowed to attend only to the committed history and to preceding chunks within the active window that satisfy the chunk-wise causal constraint. The latent state of chunk \(G_m\) is updated as
\begin{equation}
z^{s+1}_{G_m}=z^{s}_{G_m}-\Delta\tau^{s}_{m}\,
\left[v_{\theta}(z^s,\tau^s,c)\right]_{G_m},
\qquad \Delta\tau^s_m\le 0.
\end{equation}

Once a chunk reaches the data endpoint, its latent is finalized and its Transformer key-value pairs are committed to a rolling KV cache. Owing to the chunk-wise causal attention structure, committed chunks will never be modified by subsequent denoising steps, allowing their cached representations to be safely reused without recomputation. Consequently, MiniWorld supports arbitrarily long video generation while maintaining a bounded active attention window.

\noindent{\sffamily\bfseries Structured Rolling KV Cache.}
MiniWorld partitions the inference state into committed history and an active denoising window. The committed history consists of a persistent sink anchor together with a fixed-length FIFO cache, while future chunks remain inside the active window and are progressively denoised according to the chunk-wise autoregressive schedule.

As generation proceeds, newly completed chunks are committed to the cache, while newly initialized noisy chunks continuously enter the active window. Once the cache exceeds its predefined capacity, the oldest cached chunks are discarded. The initial clean context is permanently retained as a sink anchor, whereas only subsequent history chunks participate in the FIFO eviction process. Throughout inference, the active window therefore maintains a fixed computational cost independent of the generated sequence length.

\noindent{\sffamily\bfseries RoPE Re-shifting.}
Although the KV cache has a fixed capacity, newly generated chunks continuously advance the temporal positions. Directly appending new chunks would therefore require extrapolating Rotary Position Embeddings (RoPE) beyond the position range observed during training. To avoid this issue, after each cache update we apply RoPE re-shifting to the retained keys by subtracting the temporal offset introduced by the evicted chunks, effectively re-indexing the sliding window from the beginning. Since values are independent of positional rotations, they remain unchanged. When a sink anchor is present, it always occupies the first position and is excluded from the re-shifting operation, while the remaining cached chunks are shifted accordingly. This procedure keeps both queries and cached keys within the positional range seen during training without altering their relative temporal relationships.

\noindent{\sffamily\bfseries Pipelined Denoising.}
MiniWorld allows multiple future chunks to remain at different diffusion timesteps simultaneously. Consequently, several chunks can be denoised in parallel within the active window, forming a pipelined inference process. Since the number of denoising updates available to each chunk depends on the autoregressive stride and the size of the active window, the sampler adaptively allocates the timestep interval \(\Delta\tau\) such that every chunk reaches the data endpoint before being committed to the cache.

This asynchronous denoising strategy enables a flexible trade-off between inference throughput and generation quality by adjusting the pipeline depth, while requiring no modification to the trained model parameters. Combined with chunk-wise causal attention, block-wise noise scheduling, and long-horizon training, it forms the complete streaming inference framework of MiniWorld.

\section{Experiments}

\subsection{Experimental Setup}

\noindent{\sffamily\bfseries Datasets.}
We evaluate MiniWorld on two real-world benchmarks with different control modalities: DROID~\citep{droid} and RealEstate10K (RE10K)~\citep{re10k}. DROID evaluates embodied world modeling under low-level robot actions, while RE10K evaluates camera-controlled scene prediction under camera trajectories. All videos are resized to \(240\times320\) and encoded into the latent space of the pretrained Wan2.2 VAE, which has a temporal compression ratio of \(4\times\), a spatial compression ratio of \(16\times\), and 48 latent channels.

\noindent{\sffamily\bfseries Condition Processing.}
MiniWorld uses the same latent-frame-aligned conditioning interface across domains. For DROID, each video frame is paired with a 7-dimensional robot action consisting of a 6-DoF Cartesian end-effector displacement and a gripper position. Each action dimension is normalized by percentile statistics,
\begin{equation}
a \leftarrow \frac{2(a-q_{0.01})}{q_{0.99}-q_{0.01}}-1,
\end{equation}
and clipped to \([-1,1]\). Since one latent frame corresponds to four RGB frames, four consecutive robot actions are concatenated into a 28-dimensional latent-frame action token.

For RE10K, camera intrinsics and extrinsics are converted into per-pixel ray origins and directions following the NeRF convention. Each scalar is encoded with 15 sinusoidal frequency bands, producing a 180-channel pose representation per RGB frame; four consecutive pose maps are concatenated to obtain a 720-channel spatial condition aligned with each latent frame.

\noindent{\sffamily\bfseries Training Details.}
All models are trained from scratch using the recipe in Section~\ref{sec:method}. We use a latent chunk size of 2, so each causal chunk contains two consecutive latent frames. Training proceeds in two stages. In the short-horizon stage, we first train on 21-frame clips and then continue on 46-frame clips, using a global batch size of 64 and a learning rate of \(1\times10^{-4}\) with the Muon optimizer~\citep{muon}. This stage encourages efficient learning of local action-conditioned dynamics while maximizing sample coverage.

We then enter the long-video continued-training stage, where the model is trained on 125-frame sequences before being further continued on 253-frame rollouts. We adopt the Muon optimizer with a global batch size of 8, and linearly warm up the learning rate from 0 to \(2\times10^{-5}\) at the start of long-video continued training. This stage exposes the model to significantly longer causal contexts, bringing the training distribution closer to that encountered during streaming inference.

For the noise scheduler, we use 50 discretized training-time bins with monotonic CoPP timestep compositions; timesteps are shifted using the SD3-style reparameterization~\citep{sd3}, with the shift factor automatically determined by the number of latent tokens. For timestep sampling, we use the default logit-normal distribution with \(P_{\mathrm{mean}}=0\) and \(P_{\mathrm{std}}=1\). During training, the action condition is randomly dropped with a probability of 10\% to enable classifier-free guidance at inference time. All experiments are conducted on a single server equipped with 8 GPUs using bf16 mixed-precision training.

\noindent{\sffamily\bfseries Model Configuration.}
MiniWorld adopts a unified architecture family across model scales. All variants use the same block-causal Transformer design, conditioning interface, patch size, and MLP ratio, and differ only in depth, width, number of attention heads, and parameter count. Table~\ref{tab:model-config} summarizes the model configurations used in this study.

\begin{table}[t]
  \centering
  \caption{Model configurations. All variants use the same block-causal attention pattern, conditioning interface, patch size, and MLP ratio.}
  \label{tab:model-config}
  \small
  \setlength{\tabcolsep}{4pt}
  \begin{tabular}{lccccc}
    \toprule
    Model & Depth & Width & Heads & MLP ratio & Parameters \\
    \midrule
    MiniWorld-B & 12 & 768 & 12 & 4.0 & 0.12B \\
    MiniWorld-L & 24 & 1024 & 16 & 4.0 & 0.39B \\
    MiniWorld-0.5B & 28 & 1152 & 16 & 4.0 & 0.55B \\
    MiniWorld-1B & 28 & 1536 & 12 & 4.0 & 1B \\
    MiniWorld-3B & 32 & 2560 & 20 & 4.0 & 3B \\
    \bottomrule
  \end{tabular}
\end{table}

\noindent{\sffamily\bfseries Evaluation Protocol.}
We compare MiniWorld with a bidirectional short-video baseline, which performs sliding-window prediction using a fixed 29-frame window under the same control sequence. All evaluations use a single observed frame as context. The main figures report relative scores, with the bidirectional short-video baseline normalized to 1. For error metrics such as LPIPS and depth error, we invert the ratio so that larger values consistently indicate better performance. In addition to PSNR, SSIM, and LPIPS~\citep{ssim,lpips}, we report WorldArena-style metrics~\citep{worldarena} covering appearance, dynamics, geometry, VLM-based functional quality, and fidelity. Unless otherwise specified, MiniWorld uses classifier-free guidance with scale 2 during quality evaluation. The main results use MiniWorld-1B on 50 held-out videos with 253-frame streaming rollouts.

\subsection{Main Results}

\begin{figure}[!t]
  \centering
  \begin{subfigure}[t]{0.49\linewidth}
    \centering
    \includegraphics[width=\linewidth]{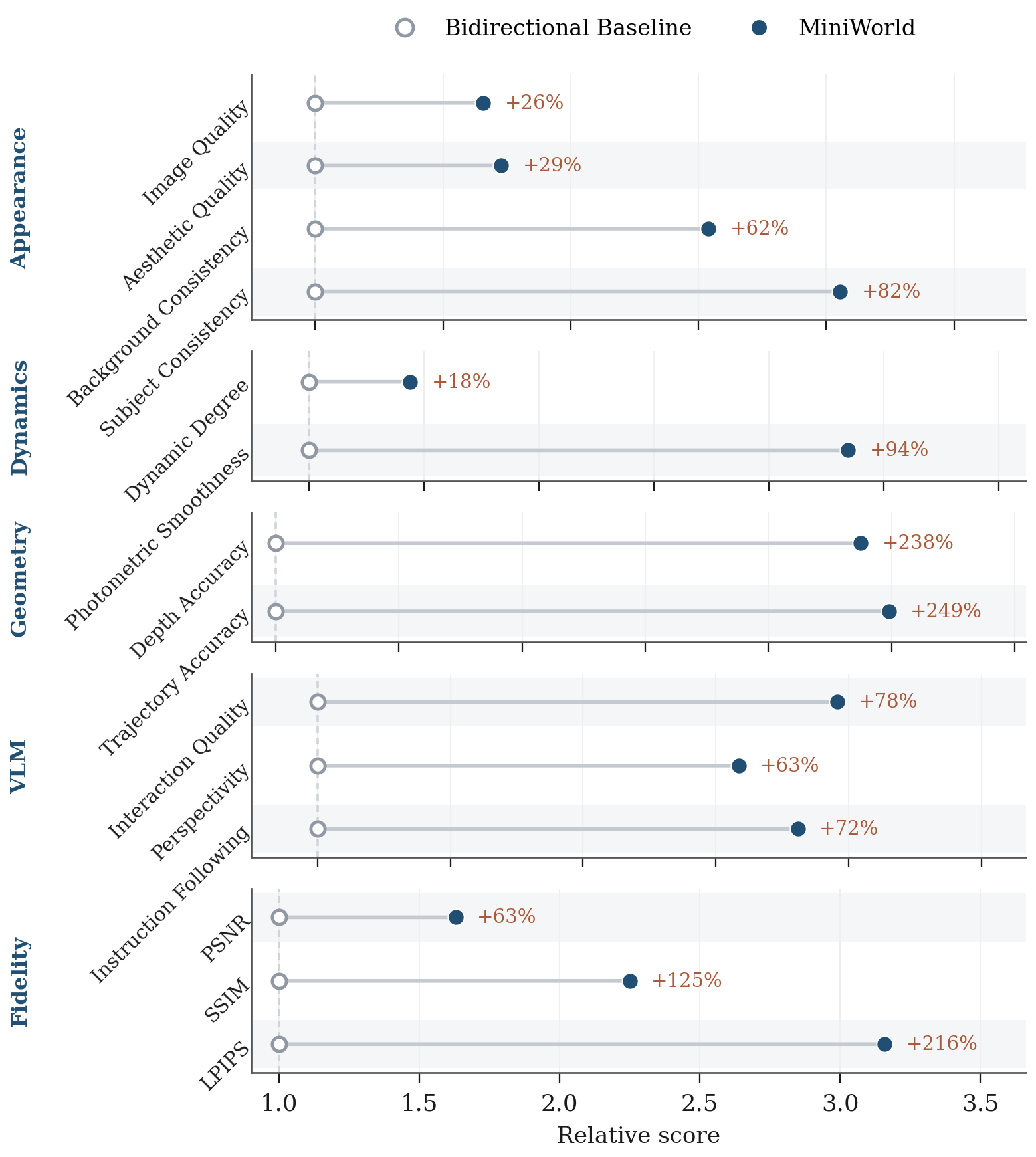}
    \caption{DROID.}
    \label{fig:main-droid}
  \end{subfigure}
  \hfill
  \begin{subfigure}[t]{0.49\linewidth}
    \centering
    \includegraphics[width=\linewidth]{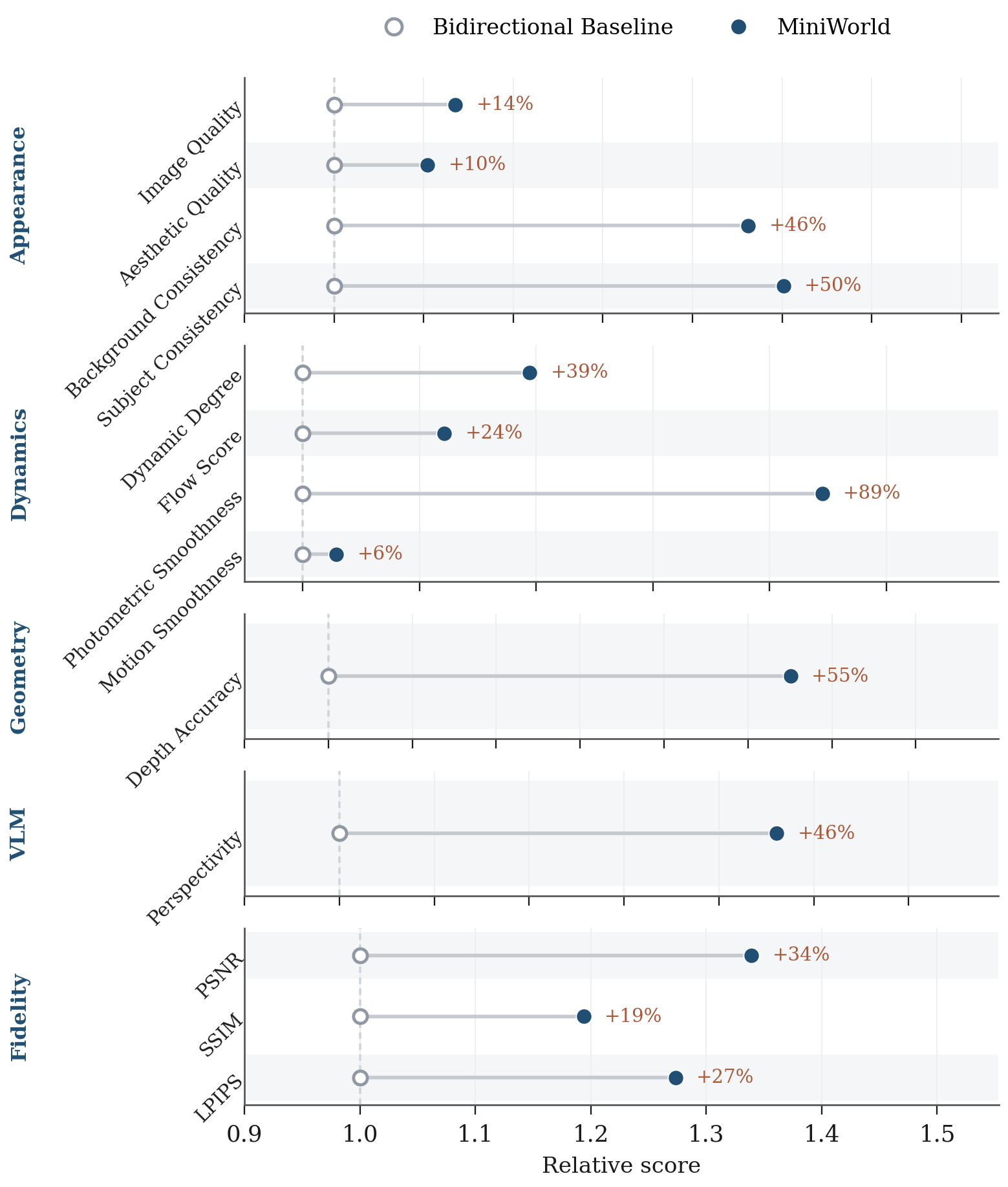}
    \caption{RealEstate10K.}
    \label{fig:main-re10k}
  \end{subfigure}
  \caption{Main results on DROID and RealEstate10K. The bidirectional short-video baseline is normalized to 1. For error metrics, the relative ratio is inverted so that larger values indicate better performance.}
  \label{fig:main-results}
\end{figure}

Figure~\ref{fig:main-droid} reports the DROID comparison. MiniWorld improves nearly all displayed metrics over the bidirectional short-video baseline, with especially large gains in geometry and fidelity. Trajectory Accuracy improves by 249\%, Depth Accuracy by 238\%, LPIPS by 216\%, and SSIM by 125\%. Appearance metrics improve consistently by 26\% to 82\%, and VLM judge scores improve by 63\% to 78\%. These gains indicate that MiniWorld does not merely sharpen individual frames: it better preserves robot-object geometry, maintains temporally coherent manipulation dynamics, and produces behavior that is more consistent with the conditioning signal.

Figure~\ref{fig:main-re10k} reports the corresponding RE10K results. The improvements are more moderate than on DROID but remain broad across metric groups: Photometric Smoothness improves by 89\%, Depth Accuracy by 55\%, Subject Consistency by 50\%, Background Consistency by 46\%, and Perspectivity by 46\%. Standard fidelity metrics also improve, with PSNR, SSIM, and LPIPS gaining 34\%, 19\%, and 27\%, respectively. This cross-domain trend is important because RE10K removes robot interaction and instead stresses camera-conditioned geometry. The same streaming architecture therefore transfers from action-conditioned manipulation to camera-controlled scene prediction without relying on a dataset-specific evaluation artifact.

\subsection{Ablations}

\begin{figure}[!t]
  \centering
  \includegraphics[width=\linewidth]{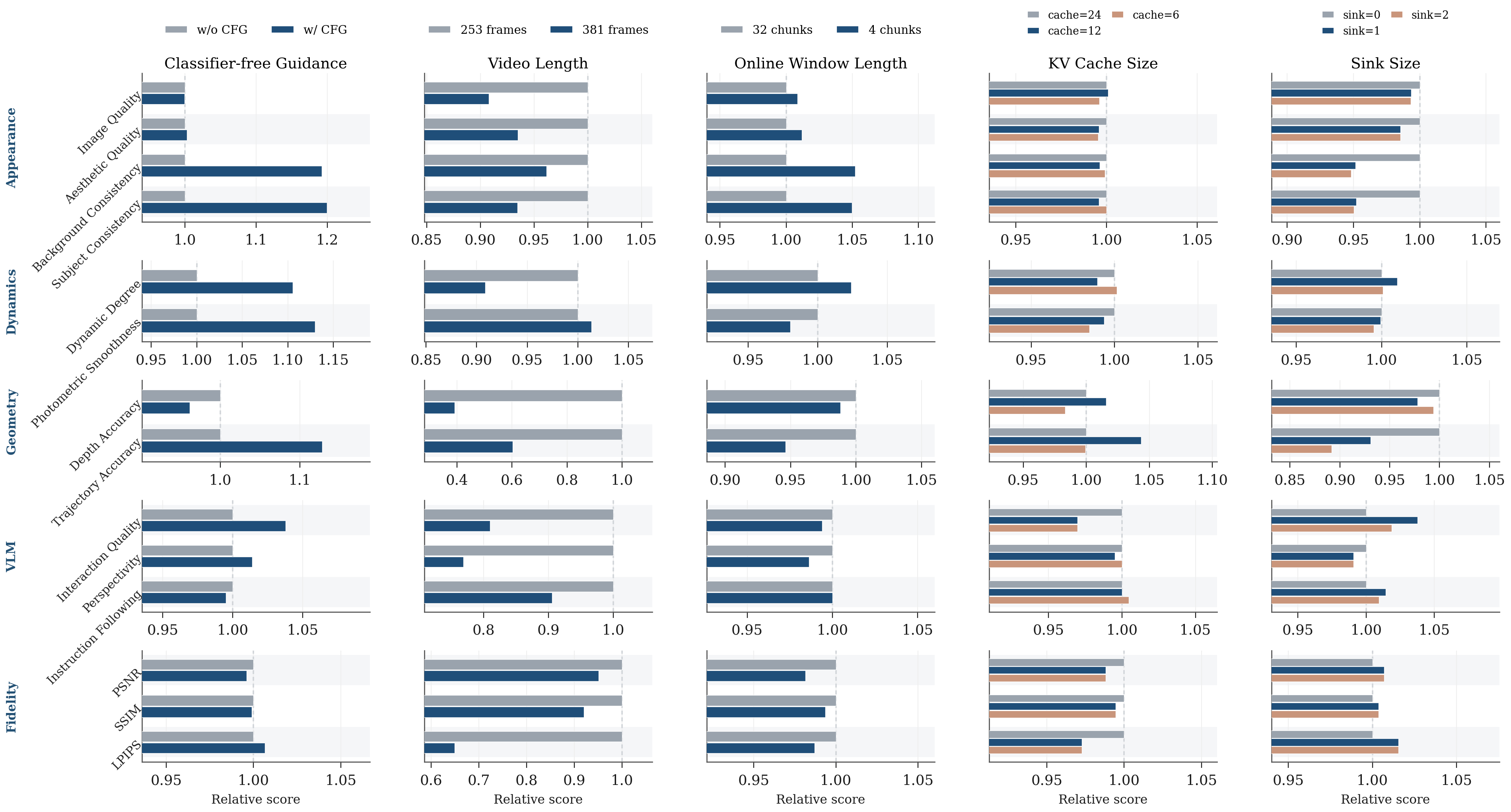}
  \caption{Quality ablations on DROID. Each column compares one inference factor while keeping the remaining settings fixed; the left setting in each column is normalized to 1.}
  \label{fig:ablation-quality}
\end{figure}

We next ablate the main inference choices on DROID. Figure~\ref{fig:ablation-quality} compares paired settings by normalizing the left setting in each column to 1, and groups metrics into Appearance, Dynamics, Geometry, VLM, and Fidelity.

First, classifier-free guidance (CFG) improves quality without materially changing frame-level fidelity. Enabling CFG improves the average displayed metric by about 5.5\%, with the clearest gains in Appearance (+9.9\%) and Dynamics (+11.8\%), while Fidelity remains nearly unchanged. This suggests that guidance acts as a clean quality knob rather than being responsible for the efficiency results.

Second, increasing the generated video length from 253 to 381 frames preserves the core visual and temporal behavior but exposes the remaining long-horizon failure modes. Appearance and Dynamics retain approximately 94\% and 96\% of the shorter-rollout score, respectively, whereas Geometry, VLM quality, and Fidelity degrade more noticeably. Thus MiniWorld remains stable over substantially longer rollouts, but geometric accuracy and perceptual fidelity remain the most sensitive dimensions as prediction errors accumulate.

Third, reducing the online denoising window from a 32-chunk full window to an 8 in-flight window with KV cache preserves nearly the same quality. The average relative score across all displayed categories is approximately unchanged: Appearance slightly improves, Dynamics remains effectively stable, and Geometry, VLM, and Fidelity stay within a few percent of the full-window setting. This result demonstrates that a small online window can match the quality of full-window inference by committing completed chunks to the rolling KV cache.

Fourth, we study whether the model requires a large retained history once the online denoising window is fixed. Using MiniWorld-1B on DROID with a 64-latent-frame rollout (\(T=64\)), eight in-flight chunks, and one resident sink frame, we vary the KV-cache size from 24 chunks to 12 and 6 chunks. The trend is largely flat: reducing the cache to 12 chunks is nearly indistinguishable from using 24 chunks, while a 6-chunk cache introduces only small single-digit changes. Appearance and Dynamics remain within approximately 1\% of the 24-chunk reference, and Geometry is similarly stable, with only minor changes in Depth Accuracy and comparable Trajectory Accuracy. These results suggest that, for this horizon, MiniWorld does not rely on retaining the full 24-chunk history once the active denoising window and sink anchor are fixed.

Finally, we ablate the number of resident sink frames using MiniWorld-1B on DROID with \(T=64\) and a 24-chunk KV cache. Compared with the no-sink setting, adding one or two sink frames is mostly neutral rather than uniformly beneficial. Appearance and Dynamics decrease slightly, while VLM-based and semantic-consistency metrics improve modestly; Trajectory Accuracy decreases mildly as the sink size increases. We therefore interpret sink anchors as a memory-stabilization mechanism that can preserve global context without collapsing quality, rather than as the primary source of the DROID quality gains. The dominant improvement instead comes from the KV-cache streaming design itself.

\subsection{Analysis}

\begin{figure}[!t]
  \centering
  \begin{subfigure}[t]{0.49\linewidth}
    \centering
    \includegraphics[width=\linewidth]{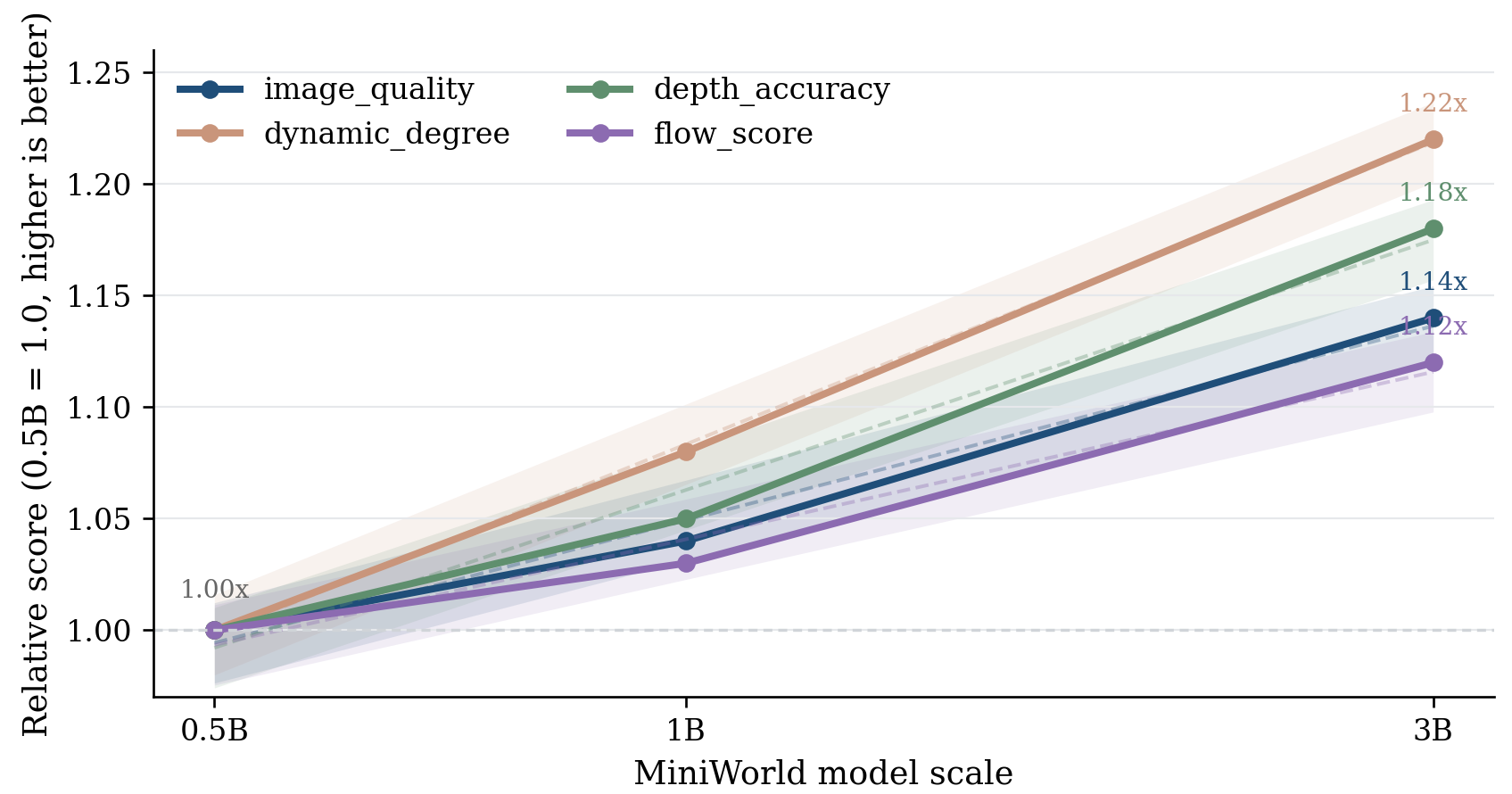}
    \caption{Model scaling on RE10K. Scores are normalized to the 0.5B model, and higher is better.}
    \label{fig:re10k-scaling}
  \end{subfigure}
  \hfill
  \begin{subfigure}[t]{0.49\linewidth}
    \centering
    \includegraphics[width=\linewidth]{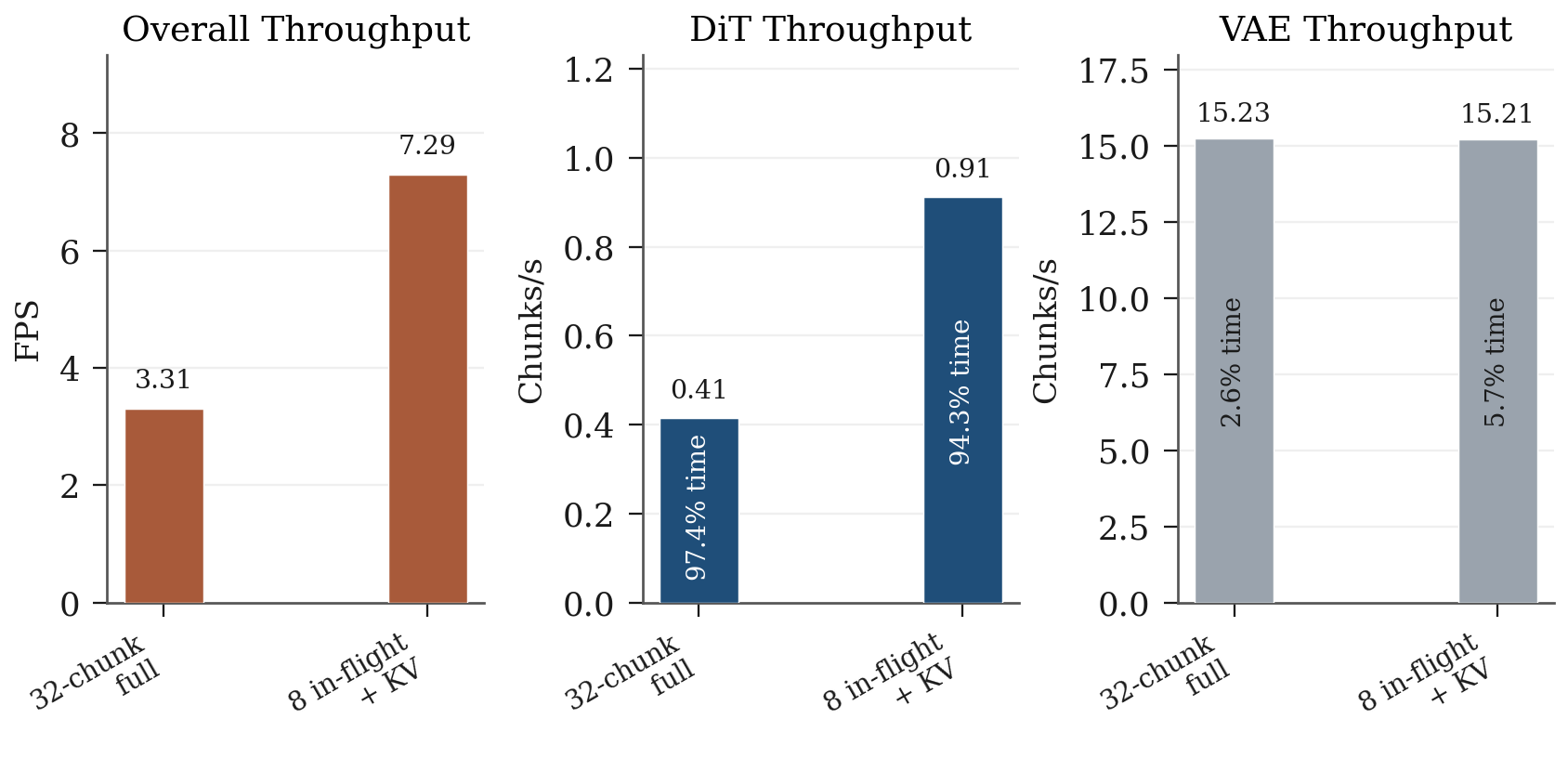}
    \caption{Streaming throughput on DROID. An 8 in-flight window with KV cache more than doubles output FPS by reducing DiT computation.}
    \label{fig:ablation-throughput}
  \end{subfigure}
  \caption{Scaling and throughput analyses of MiniWorld.}
  \label{fig:scaling-throughput}
\end{figure}

\noindent{\sffamily\bfseries Scaling Behavior.}
Figure~\ref{fig:re10k-scaling} compares MiniWorld variants on RE10K under identical training and evaluation settings and a fixed inference horizon, with all scores normalized to the 0.5B model. Performance improves monotonically from 0.5B to 1B and 3B across image quality, dynamic degree, depth accuracy, and flow score, indicating that the architecture benefits consistently from increased capacity rather than trading one aspect of generation quality for another. At 3B parameters, MiniWorld improves dynamic degree by 22\%, depth accuracy by 18\%, image quality by 14\%, and flow score by 12\% over the 0.5B model. The larger gains in dynamics and geometry show that scaling primarily strengthens motion modeling and 3D consistency, while still delivering clear improvements in appearance and flow quality.

\noindent{\sffamily\bfseries Throughput Analysis.}
Finally, we measure streaming throughput using a dedicated DROID benchmark on the 1B model. To isolate system throughput from guidance overhead, the benchmark disables classifier-free guidance (\(\mathrm{cfg}=1\)), runs 3 warmup clips and 20 timed clips per setting, and reports steady-state timing after the first generated chunk.

As shown in Figure~\ref{fig:ablation-throughput}, replacing the 32-chunk full window with an 8 in-flight window plus KV cache increases steady output throughput from 3.31 FPS to 7.29 FPS, corresponding to a 2.20\(\times\) speedup. The decomposition shows that this gain comes from reducing DiT computation: DiT throughput increases from 0.41 to 0.91 chunks/s, while VAE throughput remains around 15.2 chunks/s in both settings. Consistently, DiT accounts for 97.4\% of per-chunk compute in the full-window setting and 94.3\% with KV cache, whereas the VAE contributes only 2.6\% and 5.7\%. Therefore, the VAE is not the inference bottleneck; the efficiency improvement is achieved by bounding the online DiT attention window and reusing committed history through the rolling KV cache.

The same benchmark also shows a substantial improvement in streaming responsiveness. The first generated chunk latency decreases from 74.0 seconds to 4.86 seconds, a 15.2\(\times\) reduction. Together with the quality ablations, these results demonstrate that MiniWorld improves throughput and latency without collapsing long-horizon generation quality.

\section{Discussion}

\noindent{\sffamily\bfseries Positioning.}
MiniWorld demonstrates that a streaming video world model can be trained from scratch using an accessible and reproducible recipe under modest computational budgets. Rather than adapting a large bidirectional video generator into a causal predictor, MiniWorld directly optimizes streaming autoregressive prediction through block-causal attention, chunk-wise diffusion forcing, and train-test-aligned inference. Despite its simplicity, this formulation produces stable long-horizon rollouts across multiple domains while remaining practical for researchers without large-scale infrastructure.

Rather than pursuing state-of-the-art generation quality, MiniWorld is intended as a transparent baseline for video world modeling. While large bidirectional video generation models remain preferable when maximizing perceptual quality, MiniWorld emphasizes reproducibility and deployment consistency, making it well suited for studying streaming world models in academic research.

We hope MiniWorld provides a practical foundation for investigating problems specific to streaming world models, including temporal memory, causal representation learning, long-horizon error accumulation, and train-test-aligned optimization, while lowering the engineering barrier to developing and evaluating new algorithmic ideas.

\noindent{\sffamily\bfseries Limitations.}
Despite these encouraging results, several limitations remain. Our experiments are conducted at relatively modest model and data scales compared with frontier video foundation models, and the evaluated domains cover only a limited range of world dynamics. Although long-horizon drift is substantially reduced, prediction errors still accumulate over extended rollouts, particularly in complex interactive scenarios. Consequently, MiniWorld should be viewed as a reproducible baseline rather than the performance ceiling of streaming world models.

\noindent{\sffamily\bfseries Future Directions.}
We believe MiniWorld can be further extended along several promising directions:
\begin{itemize}[leftmargin=*]
  \item \emph{Data scaling and curation.} Building more capable world models will require both broader data coverage and higher data quality. Future efforts may expand video-action corpora with game, simulation, and real-world data, while developing world-model-centric data pipelines for quality filtering, curation, and agent-driven training--collection feedback loops.
  \item \emph{Train--test alignment.} Reducing the discrepancy between training and inference remains a key challenge for streaming world models. Promising directions include supervised fine-tuning, reinforcement learning, long-horizon self-forcing objectives, and other train--test-aligned learning strategies that explicitly mitigate autoregressive drift and improve rollout robustness over extended horizons.
  \item \emph{Efficient deployment.} As streaming world models continue to scale, deployment efficiency will become increasingly critical. Promising directions include more effective memory management and history compression, as well as few-step distillation and model quantization for reducing inference latency and computational cost.
\end{itemize}
We hope MiniWorld serves as a simple, reproducible, and extensible baseline that facilitates future research on scalable, efficient, and robust streaming world models.

\section{Related Work}

\subsection{Video World Model}

World models learn predictive environment dynamics conditioned on historical observations and control signals. Early work studied compact latent dynamics for model-based reinforcement learning~\citep{worldmodels,dreamer}, whereas recent efforts extend world modeling to visually realistic and controllable interactive simulation. Representative systems include Genie 3~\citep{genie3}, Matrix-Game 2.0~\citep{matrixgame2}, Yume-1.5~\citep{yume15}, HY-World 1.5~\citep{hyworld15}, DreamDojo~\citep{dreamdojo}, Mirage 2~\citep{mirage2}, and LingBot-World 2.0~\citep{lingbotworld2}, which leverage large-scale video corpora, foundation models, and action-conditioned generation to model controllable environment dynamics for robotics, autonomous driving, embodied agents, and interactive world creation.

Compared with these increasingly capable and often large-scale world model systems, MiniWorld focuses on a complementary direction: providing a lightweight and fully reproducible recipe for training streaming video world models from scratch under modest computational resources, thereby lowering the barrier for broader architectural and algorithmic innovation.

\subsection{Autoregressive Video Diffusion}

Recent work has explored combining diffusion models with autoregressive generation to support streaming video generation. Diffusion Forcing~\citep{df} introduces independently corrupted tokens for sequence modeling, while DFoT~\citep{dfot} extends this formulation to video with arbitrary historical conditioning. Geometry Forcing~\citep{geometryforcing} further improves action- and camera-conditioned video generation by aligning diffusion representations with a pretrained geometric foundation model, highlighting the importance of representation learning for world modeling. AR-Diffusion~\citep{ardiff} proposes non-decreasing noise schedules to better align training with autoregressive inference. Recent systems such as MAGI-1~\citep{magi} and SkyReels-V2~\citep{skyreelsv2} further demonstrate that end-to-end autoregressive diffusion can scale to long-context video generation.

A parallel and especially active direction adapts pretrained bidirectional video diffusion models into causal generators through causal attention, KV caching, rolling-window denoising, or distillation, including Ca2-VDM~\citep{ca2vdm}, Rolling Forcing~\citep{rolling}, Causal Forcing~\citep{causalforcing}, Self Forcing~\citep{selfforcing}, and CausVid~\citep{causvid}, among others. Compared with these approaches, MiniWorld deliberately prioritizes simplicity and reproducibility, enabling end-to-end training from scratch using modest computational resources and substantially lowering the barrier to developing streaming video world models.

\section{Conclusion}

We presented MiniWorld, a reproducible framework that enables training streaming video world models from scratch under modest computational budgets. By combining a block-causal Video DiT, a chunk-oriented probability propagation scheduler, two-stage continued training, and streaming inference with a structured rolling KV cache, MiniWorld aligns training with deployment-time autoregressive generation while retaining the scalability and stability of latent diffusion models. Experiments across robot-action-conditioned and camera-pose-conditioned benchmarks demonstrate that MiniWorld produces stable and competitive long-horizon streaming predictions under diverse control modalities.

More importantly, MiniWorld suggests that training streaming video world models can be approached through a simple, transparent, and reproducible recipe rather than increasingly complex training pipelines. We hope MiniWorld serves as a practical baseline for the community, facilitating future advances in memory mechanisms, long-horizon generation stability, richer open-source action-conditioned datasets, and more robust and reproducible training pipelines for scalable world models.

\clearpage

\bibliographystyle{plainnat}
\setlength{\bibhang}{0pt}
\setlength\bibindent{0pt}
\bibliography{main}

\end{document}